\documentclass[]{article} 
\usepackage{natbib}

\usepackage[utf8]{inputenc} 
\usepackage[T1]{fontenc}    
\usepackage{graphicx}
\usepackage{hyperref}       
\usepackage{url}            
\usepackage{booktabs}       
\usepackage{amsmath}
\usepackage{amssymb}
\usepackage{amsfonts}       
\usepackage{nicefrac}       
\usepackage{microtype}      
\usepackage{alltt}
\usepackage{listings}
\usepackage{array}
\usepackage[noline, titlenumbered, vlined]{algorithm2e}
\usepackage{lmodern}
\usepackage{caption}
\usepackage{subcaption}
\usepackage{xcolor}
\SetKwComment{Comment}{$\triangleright$\ }{}

\SetCommentSty{mycommfont}
\SetKwBlock{Repeat}{Repeat}{end}
\usepackage{geometry}

\newcommand{\secref}[1]{Section~\ref{#1}}
\renewcommand{\eqref}[1]{Equation~\ref{#1}}

\newcommand{\figref}[1]{Figure~\ref{#1}}

\newcommand{\algref}[1]{Algorithm~\ref{#1}}

\newcommand{\cut}[1]{}

\newcommand{\spd}{\fontfamily{cmr}\textsc{\small StratPD}}
\newcommand{\cspd}{\fontfamily{cmr}\textsc{\small CatStratPD}}

\newcommand{\Xnj}{${\bf X}_{\texttt{\char`\\} j}$}
\newcommand{\xnj}{$x_{\texttt{\char`\\} j}$}

\renewcommand{\xi}{x^{(i)}}

\renewcommand{\slash}{\texttt{\char`\\}}

\DeclareMathOperator{\Ex}{\mathbb{E}}

\begin{document}

\title{Technical Report: Partial Dependence through Stratification}

\author{Terence Parr\\
  University of San Francisco\\
  {\tt\small parrt@cs.usfca.edu}
  \and
  James D. Wilson \\
  University of San Francisco\\
  {\tt\small jdwilson4@usfca.edu} 
}

\cut{ 
\author{Terence Parr \and James D. Wilson}
\institute{Terence Parr \at
  University of San Francisco\\
  \email{{\tt parrt@cs.usfca.edu}}
  \and
  James D. Wilson \at
  University of San Francisco\\
  \email{{\tt jdwilson4@usfca.edu}} 
}
}


\maketitle

\begin{abstract}
Partial dependence curves (FPD) introduced by \citet{PDP}, are an important model interpretation tool, but are often not accessible to business analysts and scientists who typically lack the skills to choose, tune, and assess machine learning models.  It is also common for the same partial dependence algorithm on the same data to give meaningfully different curves for different models, which calls into question their precision.  Expertise is required to distinguish between model artifacts and true relationships in the data.

In this paper, we contribute methods for computing partial dependence curves, for both numerical (\spd) and categorical explanatory variables (\cspd), that work directly from training data rather than predictions of a model. Our methods provide a direct estimate of partial dependence, and rely on approximating the partial derivative of an unknown regression function without first fitting a model and then approximating its partial derivative. We investigate settings where contemporary partial dependence methods---including FPD, ALE, and SHAP methods---give biased results. Furthermore, we demonstrate that our approach works correctly on synthetic and plausibly on real data sets.  Our goal is not to argue that model-based techniques are not useful. Rather, we hope to open a new line of inquiry into nonparametric partial dependence.
\end{abstract}

\section{Introduction}

Partial dependence, the isolated effect of a specific variable or variables on the response variable, $y$, is important to researchers and practitioners in many disparate fields such as medicine, business, and the social sciences. For example, in medicine, researchers are interested in the relationship between an individual's demographics or clinical features and their susceptibility to illness. Business analysts at a car manufacturer might need to know how changes in their supply chain affect defect rates. Climate scientists are interested in how different atmospheric carbon levels affect temperature.

For an explanatory matrix, $\bf X$, with a single (column) variable, $x_1$, a plot of the $y$ against $x_1$ visualizes the marginal effect of feature $x_1$ on $y$ exactly. Given two or more features, one can similarly plot the marginal effects of each feature separately, however, the analysis is complicated by the interactions of the variables.   Variable interactions and codependencies between features result in marginal plots that do not isolate the specific contribution of a feature of interest to the response. For example, a marginal plot of sex (male/female) against body weight would likely show that, on average, men are heavier than women. While true, men are also taller than women on average, which likely accounts for most of the difference in average weight. It is unlikely that two ``identical'' people, differing only in sex, would be appreciably different in weight.  

Rather than looking directly at the data, there are several partial dependence techniques that interrogate fitted models provided by the user: Friedman's original partial dependence \citep{PDP} (which we will denote FPD), functional ANOVA \citep{fanova}, Individual Conditional Expectations (ICE) \citep{ICE}, Accumulated Local Effects (ALE) \citep{ALE}, and most recently SHAP \citep{shap}.  Model-based techniques dominate the partial dependence research literature because interpreting the output of a fitted model  has several advantages.  For example, models have a tendency to smooth over noise. Models act like analysis preprocessing steps, potentially reducing the computational burden on model-based partial dependence techniques; e.g., ALE is $O(n)$ for the $n$ records of $\bf X$. Model-based techniques are typically model-agnostic, though for efficiency, some provide model-specific optimizations, as SHAP does. Partial dependence techniques that interrogate models also provide insight into the models themselves; i.e., how variables affect model behavior.  It is also true that, in some cases, a predictive model is the primary goal so creating a suitable model is not an extra burden.

Model-based techniques do have two significant disadvantages, however. 
The first relates to their ability to tease apart the effect of codependent features because models are sometimes required to extrapolate into regions of nonexistent support or even into nonsensical observations; e.g., see discussions in \citet{ALE} and \citet{fanova}.  As we demonstrate in \secref{sec:experiments}, using synthetic and real data sets, model-based techniques can vary in their ability to isolate variable effects in practice.  Second, there are vast armies of business analysts and scientists at work that need to analyze data, in a manner akin to exploratory data analysis (EDA), that have no intention of creating a predictive model.  Either they have no need, perhaps needing only partial dependence plots, or they lack the expertise to choose, tune, and assess models (or write software at all).

\begin{figure}
\begin{center}
\includegraphics[scale=0.65]{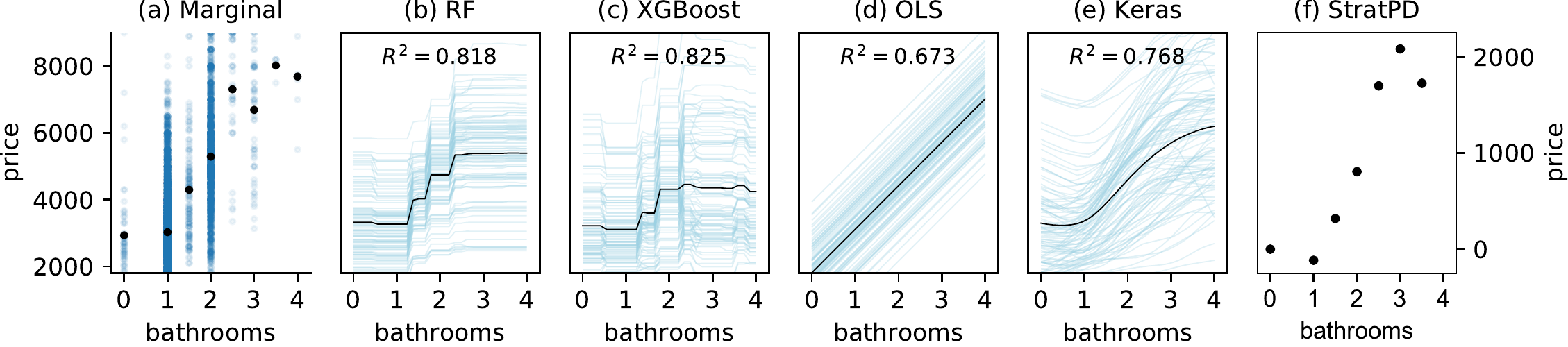}\vspace{-3mm}
\caption{\small Meaningfully different results from a single partial dependence technique, FPD/ICE, applied to the same data but different models. Plots of number of bathrooms versus rent price using New York City apartment rent data \citep{rent} with $n=10,000$ of \textasciitilde50k. (a) marginal plot, (b) plot derived from random forest, (c)  plot derived from gradient boosted machine, and (d) plot derived from ordinary least squares regression. Hyper parameters were tuned using 5-fold cross validation grid search over several hyper parameters. Keras model trained by experimentation: single hidden layer of 100 neurons, 500 epochs, batch size of 1000, batch normalization, and 30\% dropout. \spd{} gives a plausible roughly result that rent goes up linearly with the number of bathrooms. $R^2$ were computed on 20\% validation sets.\vspace{-7mm}}
\label{fig:baths_price}
\end{center}
\end{figure}

Even in the case where a machine learning practitioner is available to create a fitted model for the analyst, hazards exist. If a fitted model is unable to accurately capture the relationship between features and $y$ accurately, for whatever reason, then partial dependence does not provide any useful information to the user.  To make interpretation more challenging, there is no definition of ``accurate enough.'' Also, given an accurate fitted model, business analysts and scientists are still peering at the data through the lens of the model, which can distort partial dependence curves. Separating visual artifacts of the model from real effects present in the data requires expertise in model behavior (and optimally in the implementation of model fitting algorithms). 

Consider the combined FPD/ICE plots shown in \figref{fig:baths_price} derived from several models (random forest, gradient boosting, linear regression, deep learning) fitted to the same New York City rent data set \citep{rent}.  The subplots in \figref{fig:baths_price}(b)-(e)  present starkly different partial dependence relationships and it is unclear which, if any, is correct.  The marginal plot, (a), drawn directly from the data shows a roughly linear growth in price for a rise in the number of bathrooms, but this relationship is biased because of the dependence of bathrooms on other variables, such as the number of bedrooms. (e.g., five bathroom, one bedroom apartments are unlikely.)  For real data sets with codependent features, the true relationship is unknown so it is hard to evaluate the correctness of the plots. (Humans are unreliable estimators, which is why we need data analysis algorithms in the first place.) Nonetheless, having the same algorithm, operating on the same data, give meaningfully different partial dependences is undesirable and makes one question their  precision.

Experts are often able to quickly recognize model artifacts, such as the  stairstep phenomenon in \figref{fig:baths_price}(b) and (c) inherent to decision tree-based methods trying unsuccessfully to extrapolate.  In this case, though, the stairstep is more accurate than the linear relationship in (d) and (e) because the number of bathrooms is discrete (except for ``half baths'').  The point is that interpreting model-based partial dependence plots can be misleading, even for experts. 

An accurate mechanism to compute partial dependences that did not peer through fitted models would be most welcome.  Such partial dependence curves would be accessible to users, like business analysts, who lack the expertise to create suitable models. (One can imagine a spreadsheet plug-in that produced partial dependence curves.) A mechanism that did not rely on a user-provided model would also reduce the chance of plot misinterpretation due to model artifacts and could even help machine learning practitioners to choose appropriate models based upon relationships exposed in the data.

In this paper, we propose a strategy, called {\textsc{strat}ified \textsc{p}artial \textsc{d}ependence} (\spd{}), that computes partial dependences directly from training data $({\bf X}, {\bf y})$, rather than through the predictions of a fitted model, and that does not presume mutually-independent features. As an example, \figref{fig:baths_price}(f) shows the partial dependence plot computed by \spd.  \spd{} operates very much like a ``model-free'' ALE, at least for numerical variables.  Our technique is also based upon the notion of an idealized partial dependence:  integration over the partial derivative of $y$ with respect to the variable of interest for the smooth function that generated $({\bf X}, {\bf y})$. As that function is unknown, we estimate the partial derivatives from the data non-parametrically.  Colloquially, the approach examines changes in $y$ across $x_j$ while holding \xnj{} constant or nearly constant (\xnj{} denotes all variables except $x_j$).  To hold \xnj{} constant, we use a single decision tree to partition feature space, a concept used by \citet{rfimp} and \citet{RFunsup} for conditional permutation importance and observation similarity measures, respectively.  Our second contribution, \cspd{}, computes partial dependence curves for categorical variables that, unlike existing techniques, does not assume adjacent category levels are similar. Both \spd{} and \cspd{} are quadratic in $n$, in the worst case (like FPD), though \spd{} behaves linearly on real data sets.  Our prototype is currently limited to regression, isolates only single-variable partial dependence, and cannot identify interaction effects (as ICE can).  The software is available via Python package {\tt stratx} with source at {\tt github.com/parrt/stratx}, including the code to regenerate images in this paper.

We begin by describing the proposed stratification approach in \secref{sec:stratpd} then compare \spd{} to related (model-based) work in \secref{sec:related}. In \secref{sec:experiments}, we present partial dependence curves generated by \spd{} and \cspd{} on synthetic and real data sets, contrast the plots with those of existing methods, and use synthetic data to highlight possible bias in some model-based methods.

\section{Partial dependence without model predictions}\label{sec:stratpd}

In special circumstances, we know the precise effect of each feature $x_j$ on response $y$.  Assume we are given training data pair ($\bf X, y$) where ${\bf X} = [x^{(1)}, \ldots, x^{(n)}]$ is an $n \times p$ matrix whose $p$ columns represent observed features and ${\bf y}$ is the $n \times 1$ vector of responses. For any smooth function $f:\mathbb{R}^{p} \rightarrow \mathbb{R}$ that precisely maps each $\xi$ row vector to response $y^{(i)}$, ${y^{(i)}} = f(\xi)$, the partial derivative of $y$ with respect to $x_j$ gives the change in $y$ holding all other variables constant.  Integrating the partial derivative then gives the {\em idealized partial dependence}  of $y$ on $x_j$, the isolated contribution of $x_j$ to $y$:

\noindent {\bf Definition 1} The {\em idealized partial dependence} of $y$ on feature $x_j$ for continuous and smooth generator function $f:\mathbb{R}^{p} \rightarrow \mathbb{R}$ evaluated at $x_j = z$ is the cumulative sum up to $z$:\vspace{-1mm}

\begin{equation}\label{eq:pd}
\text{\it PD}_j(z) = \int_{min(x_j)}^z \frac{\partial f}{\partial x_j} dx_j
\end{equation}\vspace{-1mm}

\noindent $\text{\it PD}_j(z)$ is the value contributed to $f$ by $x_j$ at $x_j = z$ and $\text{\it PD}_j(min(x_j))=0$. The underlying generator function is unknown, and so other approaches begin by estimating $f$ with a fitted model, $\hat{f}$. Instead, we estimate the partial derivatives of the true function, ${\partial f}/{\partial x_j}$, from the raw training data then integrate to obtain $PD_j$. We note that ALE also derives partial dependence by estimating and integrating across partial derivatives (e.g., see Equation 7 in \citealt{ALE}) but does so using local changes in predictions from $\hat{f}$ rather than $f$.  The advantages of this $PD_j$ definition are that it ({\em i}) does not require a fitted model or its predictions and ({\em ii}) is insensitive to codependent features.  

The key idea is to stratify \xnj{} feature space into disjoint regions of observations where all \xnj{} variables are approximately matched across the observations in that region. Within each \xnj{} region, any fluctuations in the response variable are likely due to the variable of interest, $x_j$.  In the ideal case, the values for each \xnj{} variable within a region are identical and, because only $x_j$ is changing, $y$ changes should be attributed to $x_j$, noise, or irreducible error. (More on this shortly.) Estimates of the partial derivative within a region are computed discretely as the changes in $y$ values between unique and ordered $x_j$ positions:  $(\bar{y}^{(i+1)} - \bar{y}^{(i)})/(x_j^{(i+1)} - x_j^{(i)})$ for all $i$ in a region such that $x_j^{(i)}$ is unique and $\bar{y}^{(i)}$ is the average $y$ at unique $x_j^{(i)}$.  This amounts to performing piecewise linear regression through the region observations, one model per unique pair of $x_j$ values, and collecting the model $\beta_1$ coefficients to estimate partial derivatives. The overall partial derivative at $x_j=z$ is the average of all slopes, found in any region, whose range $[x_j^{(i)},x_j^{(i+1)})$ spans $z$.  One could apply a nonparametric method to smooth through the discontinuities at $x_j$ points within a leaf, but this has not proven necessary in practice.  The partial dependence curve points are often the result of two averaging operations, one within and one across regions, which tends to smooth the curve; e.g., see \figref{fig:interactions}(d) below. 

Stratification occurs through the use of a decision tree fit to (\Xnj, $\bf y$), whose leaves aggregate observations with equal or similar \xnj{} features. The \xnj{} features can be numerical variables or label-encoded categorical variables (assigned a unique integer). \spd{} only uses the tree for the purpose of partitioning feature space and never uses predictions from any model. See \algref{alg:StratPD} for more details.

For the stratification approach to work, decision tree leaves must satisfactorily stratify \xnj{}. If the \xnj{} observations in each region are not similar enough, the relationship between $x_j$ and $y$   is less accurate.  Regions can also become so small that even the $x_j$ values become equal, leaving a single unique $x_j$ observation in a leaf. Without a change in $x_j$, no partial derivative estimate is possible and these nonsupporting observations must be  ignored (e.g., the two leftmost points in \figref{fig:partitioning}(a)). A degenerate case occurs when identical or nearly identical $x_j$ and $x_j'$ variables exist. Stratifying $x_j$ as part of \xnj{} would also match up $x_j'$ values, leading to both exhibiting flat curves, as if the decision tree were trained on $(\bf X, y)$ not (\Xnj, $\bf y$). Our experience is that using the collection of leaves from a random forest, which restricts the number of variables available during node splitting, prevents partitioning from relying too heavily on either $x_j$ or $x_j'$. Some leaves have observations that vary in $x_j$ or $x_j'$ and partial derivatives can still be estimated.

\spd{} uses a hyper parameter, {\tt\small min\_samples\_leaf}, to control the minimum number of observations in each decision tree leaf.  Generally speaking, smaller values lead to more confidence that fluctuations in $y$ are due solely to $x_j$, but more observations per leaf 
prevent \spd{} from missing nonlinearities and make it less susceptible to noise.  As the leaf size grows, however, one risks introducing contributions from \xnj{} into the relationship between $x_j$ and $y$. At the extreme, the decision tree would consist of a single leaf node containing all observations, leading to a marginal not partial dependence curve.

\spd{} uses another hyper parameter called {\tt\small min\_slopes\_per\_x} to ignore any partial derivatives estimated with too few observations.  Dropping uncertain partial derivatives greatly improves accuracy and stability. Partial dependences computed by integrating over local partial derivatives are highly sensitive to partial derivatives computed at the left edge of any $x_j$'s range because imprecision at the left edge affects the entire curve.  This presents a problem when there are few samples with $x_j$ values at the extreme left (see, for example, the $x_j$ histogram of \figref{fig:yearmade}(d)).  Fortunately, sensible defaults for \spd{} (10 observations and 5 slopes) work well in most cases and  were used to generate all plots in this paper.

For categorical explanatory variables, \cspd{} uses the same stratification approach, but cannot apply regression of $y$ to non-ordinal, categorical $x_j$. Instead, \cspd{} groups leaf observations by category and computes the average response per category in each leaf. Consider a single leaf and its  $p$-dimensional average response vector $\bar{\bf y}$. Then choose a random reference category, {\it refcat}, and subtract that category's  average value from $\bar{\bf y}$ to get a vector of relative deltas between categories: $\Delta {\bf y}$ = $\bar{\bf y} - \bar{\bf y}_{\it refcat}$. The $\Delta  {\bf y}$ vectors from all leaves are then merged via averaging, weighted by the number of observations per category, to get the overall effect of each category on the response.  The delta vectors for two leaves, $\Delta {\bf y}$ and $\Delta {\bf y}'$, can only be merged if there is at least one category in common.  \cspd{} initializes a running average vector to an arbitrary starting leaf's $\Delta  {\bf y}$ and then makes multiple passes over the remaining vectors, merging any vectors with a category in common with the running average vector.  Observations associated with any remaining, unmerged leaves must be ignored. \cspd{} uses a single hyper parameter {\tt\small min\_samples\_leaf} to control stratification. Both \spd{} and \cspd{} have an optional hyper parameter called {\tt\small ntrials} (default is 1) that averages the results from multiple bootstrapped samples, which can reduce variance.

Stratification of high-cardinality categorical variables tends to create small groups of category subsets, which complicates the averaging process across groups. (Such $\Delta {\bf y}$ vectors are sparse and we use {\it NaN}s to represent missing values.) If both groups have the same reference category, merging is a simple matter of averaging the two delta vectors, where {\it mean}({\it z,NaN}) = {\it z}.  For delta vectors with different reference categories and at least one category in common, one vector is adjusted to use a randomly-selected reference category, $c$, in common: $\Delta {\bf y}' = \Delta {\bf y}' - \Delta {\bf y}_c' + \Delta {\bf y}_c$. That equation adjusts vector $\Delta {\bf y}'$ so $\Delta {\bf y}_c'=0$ then adds the corresponding value from $\Delta {\bf y}$ so $\Delta {\bf y}_c' = \Delta {\bf y}_c$, which renders the average of $\Delta {\bf y}$ and $\Delta {\bf y}'$ meaningful.  See \algref{alg:CatStratPD} for more details.

\spd{} and \cspd{} both have theoretical worst-case time complexity of $O(n^2)$ for $n$ observations. For \spd{}, stratification costs $O(p n \,log n)$, computing $y$ deltas for all observations among the leaves has linear cost, and averaging slopes across unique $x_j$ ranges is on the order of $|unique({\bf X}_j)| \times n$ or $n^2$ when all ${\bf X}_j$ are unique in the worst case. \spd{} is, thus, $O(n^2)$ in the worst case.  \cspd{} also stratifies in $O(p n \,log n)$ and computes category deltas linearly in $n$ but must make multiple passes over the $|T|$ leaves to average all possible leaf category delta vectors.  In practice, three passes is the max we have seen (for high-cardinality variables), so we can assume the number of passes is some small constant to get a tighter bound. Averaging two vectors costs $|unique({\bf X}_j)|$, so each pass requires $|T| \times |unique({\bf X}_j)|$. The number of leaves is roughly $n / {\tt\small min\_samples\_leaf}$ and, worst-case, $|unique({\bf X}_j)|=n$, meaning that merging dominates \cspd{} complexity leading to $O(n^2)$.  Experiments show that our prototype is fast enough for practical use (see \secref{sec:experiments}).

\section{Related work}\label{sec:related}

The mechanisms most closely related to \spd{} and \cspd{} are FPD, ICE, SHAP, and ALE, which all define partial dependence in terms of impact on estimated models, $\hat{f}$, rather than the unknown true function $f$. Let $x_S$ be the subset of features of interest where $S \subset F = \{1, 2, .., p\}$. \citet{PDP} defines the partial dependence as an expectation conditioned on the remaining variables:\vspace{-2mm}

\begin{equation}
{\it FPD}_S(x_S) = \Ex[\hat{f}(x_S,{\bf X}_{\slash S})],
\end{equation}\vspace{-3mm}

\noindent where the expectation can be estimated by $\frac{1}{n} \sum_{i=1}^n \hat{f}(x_S, x_{\slash S}^{(i)})$.
The Individual Conditional Expectation (ICE) plot \citep{ICE} estimates the partial dependence of the prediction $\hat{f}$ on $x_S$, or single variable $x_j$, across individual observations. ICE produces a curve from the fitted model over all values of $x_j$ while holding \xnj{} fixed: $\hat{f}^{(i)}_j = \widehat{f}(\{x_j^{(k)}\}_{k = 1}^n, x_{\slash j}^{(i)})$;  the FPD curve for  $x_j$ is the average over all $x_j$ ICE curves. The motivation for ICE is to identify variable interactions that average out in the FPD curve. 


The SHAP method from \citet{shap} has roots in {\em Shapley regression values} \citep{shapley-regression} and calculates the average marginal effect of adding $x_j$ to models, $\hat{f}_S$, trained on all possible subsets of features:
\vspace{-1mm}

\begin{equation}\label{eq:shap}
\phi_j(\hat{f},x_F) = \sum_{S \subseteq F \slash \{j\}}\
\frac{|S|!(|F|-|S|-1)!}{|F|!}\
 [ \hat{f}_{S \cup \{j\}}(x_{S \cup \{j\}}) - \hat{f}_S(x_S) ]
\end{equation}\vspace{-1mm}

To avoid training a combinatorial explosion of models with the various feature subsets, $\hat{f}_S(x_S)$, SHAP reuses a single model fitted to $(\bf X, y)$ by running simplified feature vectors into the model. As  \citet{manyshap} describes, there are many possible implementations for simplified vectors. One is to replace ``missing''  features with their expected value or some other baseline vector (BShap in \citealt{manyshap}). SHAP uses a more general approach (``interventional'' mode) that approximates $\hat{f}_S(x_S)$ with $\Ex[\hat{f}(x_{S},{\bf X}'_{\slash S}) | {\bf X'}_S = x_S]$ where ${\bf X}'$ is called the {\em background set} and users can pass in, for example, a single vector with ${\bf X}_{\slash S}$ column averages or even the entire training set, $\bf X$, which is what we will assume (and is called CES($\hat{D}$) in \citealt{manyshap} where $\hat{D}={\bf X}$). To further reduce computation costs, SHAP users typically explain a small subsample of the data set, but with potentially a commensurate reduction in the explanatory resolution of the underlying population. SHAP has model-type-dependent optimizations for linear regression, deep learning, and decision-tree based models.

For efficiency, SHAP approximates $\Ex[\hat{f}(x_{S},{\bf X}_{\slash S}) | {\bf X}_S = x_S]$ with $\Ex[\hat{f}(x_{S},{\bf X}_{\slash S})]$, which assumes feature independence. \citep{janzing2019feature} argues that ``{\em unconditional} [as implemented] rather than {\em conditional} [as defined] expectations provide the right notion of dropping features.''  But, using the unconditional expectation makes the inner difference of \eqref{eq:shap} a function of codependency-sensitive FPDs:

\vspace{-4mm}\begin{equation}
\begin{split}
\hat{f}_{S \cup \{j\}}(x_{S \cup \{j\}}) - \hat{f}_S(x_S) & = \Ex[\hat{f}(x_{S \cup \{j\}},{\bf X}_{\slash (S \cup \{j\})})] - \Ex[\hat{f}(x_{S},{\bf X}_{\slash S})]\\
 & = \text{\it FPD}_{S \cup \{j\}}({\bf x}) - \text{\it FPD}_{S}({\bf x})
\end{split}
\end{equation}\vspace{-2mm}

If the individual contributions are potentially biased, averaging the contributions of many such feature permutations might not lead to an accurate partial dependence.  Even if the conditional expectation is used, \citet{manyshap} points out that SHAP is sensitive to the sparsity of $\bf X$ because condition ${\bf X}_S = x_S$ will find few or no training records with the exact $x_S$ values of some input vector.


The goal of ALE \citep{ALE} is to overcome the bias in previous model-based techniques arising from extrapolations of $\hat{f}$ far outside the support of the training data in the presence of codependent variables.   ALE  partitions range $[\text{min}({\bf X}_j) \,..\, \text{max}({\bf X}_j)]$ for variable $x_j$ into $K$ bins and estimates the ``uncentered main effect'' (Equation 15) at $x_j = z$ as the cumulative sum of the partial derivatives for all bins up to the bin containing $z$. ALE estimates the partial derivative of $\hat{f}$ at $x_j=z$ as $\Ex[\hat{f}(b_k, {\bf X}_{\slash j}) - \hat{f}(b_{k-1}, {\bf X}_{\slash j}) \, | \, x_j \in (b_{k-1},b_k]]$ for bin $b_k$ that contains $z$. They also extend ALE to two variables by partitioning feature space into $K^2$ rectangular bins and computing the  second-order finite difference of $\hat{f}$ with respect to the two variables for each bin. 

Another related technique that integrates over partial derivatives to measure $x_j$ effects is called Integrated Gradients (IG) from \citet{intgrad}. Given a single input vector, $\bf x$, to a deep learning classifier, IG integrates over the gradient of the model output function $\hat{f}$ at points along the path from a baseline vector, $\bf x'$, to $\bf x$. IG can be seen as computing the partial dependence of $\hat{f}$ at a single $\bf x$, but using multiple $\bf x$ vectors would yield an $x_j$ partial dependence curve (relative to a baseline $\bf x'$).

\spd{} is like a ``model-free'' version of ALE in that \spd{} also defines partial dependence as the cumulative sum of partial derivatives, but we estimate derivatives using response values directly rather than $\hat{f}$ predictions.  An advantage to estimating partial dependence via fitted models is that $\hat{f}$ removes variability from the potentially noisy response values, $\bf y$. However, in practice, this requires expertise to choose and tune an appropriate model for a data set.  The fact that different models can lead to meaningfully different curves for the same data can lead to misinterpretation.  Also, expertise is often required to distinguish between model artifacts and interesting visual phenomena arising from the data.

\begin{figure}[!htbp]
\begin{center}
\includegraphics[scale=.4]{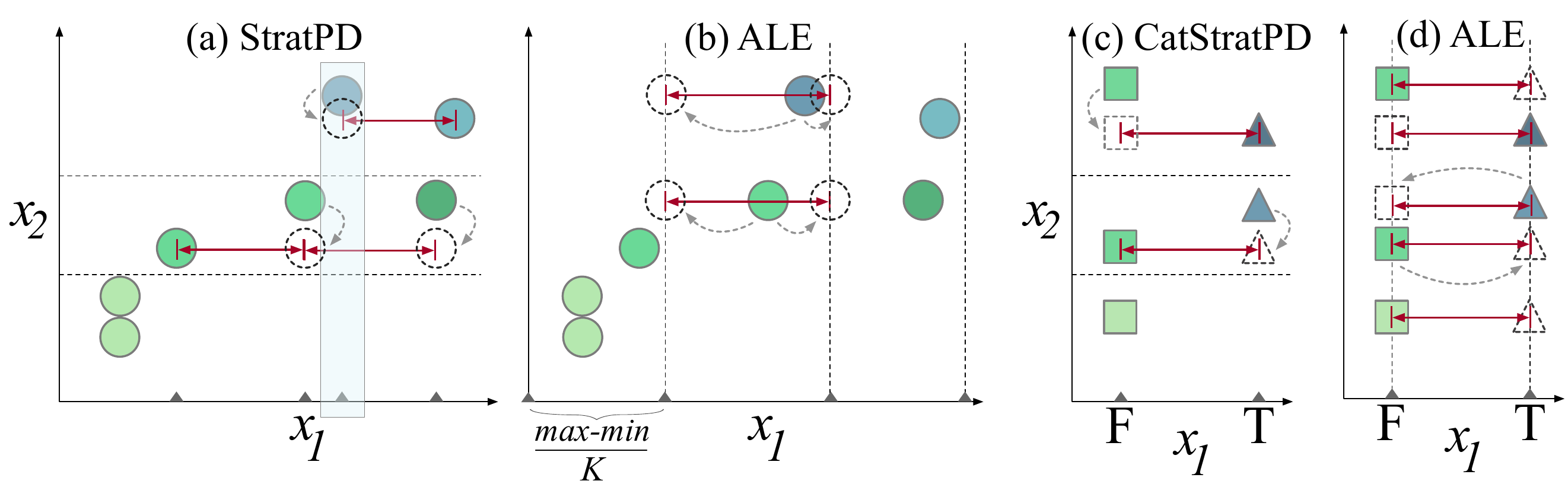}\vspace{-3mm}
\caption{\small Comparison of \spd, \cspd{} and ALE for $p=2$ with continuous $x_2$ and continuous $x_1$ in (a), (b) and  categorical $x_1$ in (c), (d).  F=false and T=true. Vertical dashed lines are ALE bin edges and horizontal dashed lines are regions partitioned by a decision tree fit to (\Xnj, ${\bf y}$). Whereas ALE shifts $x_1$ observations to bin edges holding $x_2$ exactly constant (and asks for predictions at those points), \spd{} assumes $x_2$ values are the same and uses known $y$ values. The small wedges on $x_1$ axis indicate partial dependence curve points.  The shades of green and blue indicate $y$ values. The leftmost two observations in (a) and the lowermost observation in (c) are ignored as finite differences are not defined. There is no curve value for the rightmost $x_1$ value in (a) due to forward differencing.}
\label{fig:partitioning}
\end{center}
\end{figure}

ALE partitions $x_j$ into bins then fixes \xnj{} as it shifts $x_j$ to bin edges to compute finite differences, as depicted in \figref{fig:partitioning}(b) for $p=2$. The wedges on the $x_1$ axis indicate the $x_1$ points of the computed partial dependence curve.  \spd{} partitions \xnj{} into regions of (hopefully) similar observations and computes finite differences between the average $y$ values at unique $x_j$ values in each region, as depicted in \figref{fig:partitioning}(a).  The leftmost two observations are ignored as there is no change in $x_1$ in that leaf. The shaded area illustrates that the partial derivative at any $x_j=z$ is the average of all derivatives spanning $z$ across \xnj{} regions.  \spd{} assumes all points within a region are identical in \xnj{}, effectively projecting points in \xnj{} space onto a hyperplane if they are not.  ALE shifts $x_j$ values in a small neighborhood and \spd{} depends on a suitable {\tt\small min\_samples\_leaf}  hyper parameter to prevent \xnj{} points in a regions from becoming too dissimilar.  \spd{} automatically generates more curve points in areas of high $x_j$ density, but ALE is more efficient.

Using decision trees for the purpose of partitioning feature space as \spd{} does was previously used by \citet{rfimp} to improve permutation importance for random forests by permuting $x_j$ only within the observations of a leaf. Earlier, \citet{RFunsup} defined a similarity measure between two observations according to how often they appear in the same leaf in a random forest.  Rather than partitioning  \xnj{} space like \spd, those techniques partitioned all of $x$ space.


The model-based techniques under discussion treat boolean and label-encoded categorical variables (encoded as unique integers) as numerical variables, even though there is no defined order, as depicted in \figref{fig:partitioning}(d). ALE does, however, take advantage of the lack of order to choose an $x_j$ order that reduces ``extrapolation outside the data envelope'' by measuring the similarity of \Xnj{} sample values across $x_j$ categories.  Adjacent category integers, though, could still represent the most semantically different categories, so any shift of an observation's category to extrapolate is risky. Even the smallest possible extrapolation can conjure up nonsensical observations, such as pregnant males, as we demonstrate in \figref{fig:pregnant} below where FPD, SHAP, and ALE underestimate pregnancy's effect on body weight. (For boolean $x_j$, ALE behaves like FPD.)   See \cite{stopperm} for more on the dangers of permuting features. 

In contrast, \cspd{} uses a different algorithm for categoricals and computes differences between the average $y$ for all categories within the leaf to a random reference category; see \figref{fig:partitioning}(c). These leaf delta vectors are then merged across leaves to arrive at an overall delta vector relating the relative effect of each category on $y$.  One could argue that \spd{} also extrapolates because \xnj{} could include categorical variables and not all \xnj{} records would be identical. But, our approach only assumes \xnj{} values are similar and uses known training $y$ values for finite differences, rather than asking a model to make prediction for nonsensical records, which could be wildly inaccurate. Also, the decision tree would, by definition, likely partition \xnj{} space into regions that can be treated similarly, thus, grouping semantically similar categorical levels together.

\cut{
{\bf Comparison}

${\it FPD}_j(z) = \Ex[\hat{f}(x_{j}=z,{\bf X}_{\slash j})])$

SHAP for one $S$ is $\phi_j(\hat{f},{\bf x}) = \Ex[\hat{f}(x_{S \cup \{j\}},{\bf X}_{\slash (S \cup \{j\})})  | {\bf X}_{S \cup \{j\}} = x_{S \cup \{j\}}] - \Ex[\hat{f}(x_{S},{\bf X}_{\slash S})  | {\bf X}_S = x_S ]$ for all subsets $S \subset F$ for $F = \{1, 2, .., p\}$. When $S = F \slash \{j\}$, like ALE, it becomes $\Ex[\hat{f}(x_F)] - \Ex[\hat{f}(x_{F \slash \{j\}},{\bf X}_j)]$ or $\hat{f}({\bf x}) - \text{\it FPD}_{F \slash \{j\}}(x)$ if assume independence.

ALE at $x_j=z$, it is cumsum of $\Ex[\hat{f}(b_k, {\bf X}_{\slash j}) - \hat{f}(b_{k-1}, {\bf X}_{\slash j}) \, | \, x_j \in (b_{k-1},b_k]]$ for bin $b_k$ partitioning min..max for var $j$ into $K$ intervals.

StratPD at $x_j=z$ is $\Ex[ (y^{(i_L+1)} - y^{i_L})/(x_j^{(i_L+1)} - x_j^{(i_L)}) \, | \, z \in [x_j^{(i_L+1)} - x_j^{(i_L)}) \text{ and } L \in T]$

$\Ex[ \phi_j(\hat{f},x) | {\bf X}_j = z] = FPD_j(x) - \bar{y}$ if features independent. SHAP's implementation approximates $\hat{f}$ trained on just $S$ features, $\hat{f}_S(x_S)$, with $\Ex[\hat{f}(x_{S},{\bf X}_{\bar{S}})]$ by assuming independent features.
}

\section{Experimental results}\label{sec:experiments} 

In this section, we demonstrate experimentally that \spd{} and \cspd{} compute accurate partial dependence curves for synthetic data and plausible results for a real data set. Experiments also provide evidence that existing model-based techniques can provide meaningfully-biased curves. We begin by comparing the partial dependence curves from popular techniques on synthetic data with complex interactions.\footnote{All simulations in this section were run on a 4.0 Ghz 32G RAM machine running OS X 10.13.6 with SHAP 0.34, scikit-learn 0.21.3, XGBoost 0.90, TensorFlow 2.1.0, and Python 3.7.4; ALEPlot 1.1 and R 3.6.3. A single random seed was used across simulations for graph reproducibility purposes.}

\begin{figure}[!htbp]
\begin{center}
\includegraphics[scale=0.45]{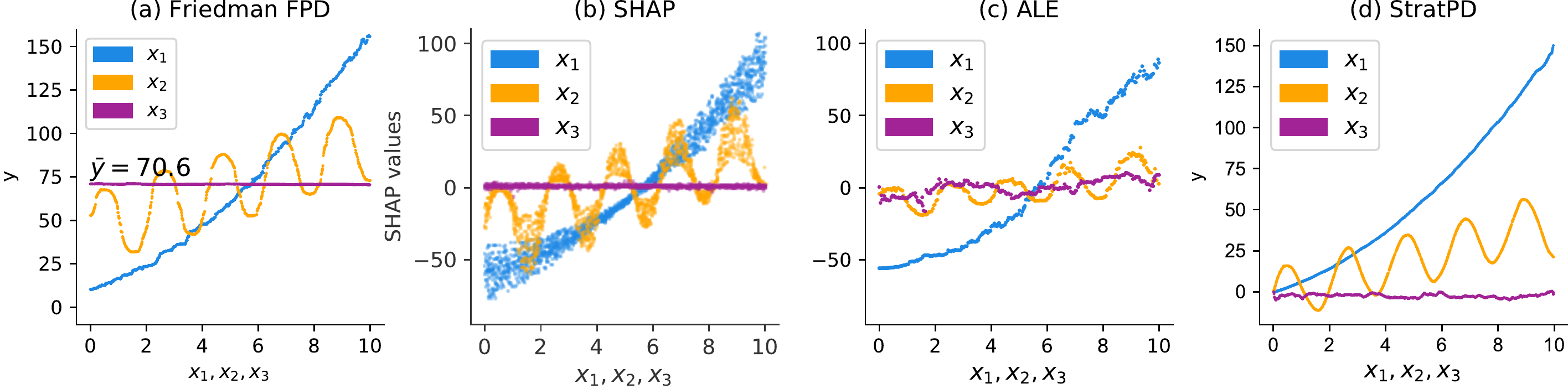}\vspace{-2mm}
\caption{\small  Partial dependence plots of $n=2000$ data generated from noiseless $y = x_1^2 + x_1 x_2 + 5 x_1 sin(3 x_2) + 10$ where $x_1,x_2,x_3 \sim U(0,10)$ and $x_3$ does not affect $y$. The model is a random forest with 10 trees trained on all data ($R^2=0.997$). SHAP used all $\bf X$ as background data and ALE used $K=300$. The curves were generated from same 2000 data points that the model was trained on.}
\label{fig:interactions}
\end{center}
\end{figure}

\figref{fig:interactions} illustrates that FPD, SHAP, ALE, and \spd{} all suitably isolate the effect of independent individual variables on the response for noiseless data generated via: $y = x_1^2 + x_1 x_2 + 5 x_1 sin(3 x_2) + 10$ for $x_1,x_2,x_3 \sim U(0,10)$ where $x_3$ does not affect $y$. The shapes of the curves for all techniques look similar except that \spd{} starts all curves at $y=0$ (as could the others). SHAP's curves have the advantage that they indicate the presence of variable interactions. To our eye, \spd's curves are smoothest despite not having access to model predictions.

\begin{figure}[!htbp]
\begin{center}
\includegraphics[scale=0.45]{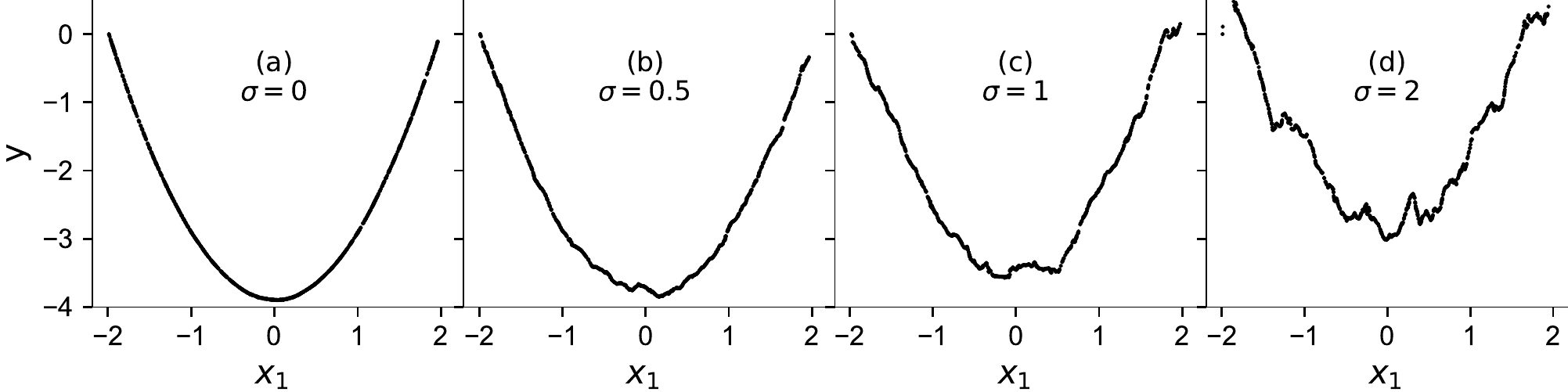}\vspace{-2mm}\caption{\small The effective noise on \spd{} for $y = x_1^2 + x_1 + 10 + \epsilon$ where $x_1,x_2 \sim U(-2,2)$, $\epsilon \sim N(0,\sigma)$ with $\sigma \in \{0,0.5,1,2\}$. The 95\% interval for amplitude of the noise in (d) for $\sigma=2$ is the same as the signal.}
\label{fig:noise}
\end{center}
\end{figure}

Models have a tendency to smooth out noise and a legitimate concern is that, without the benefit of a model, \spd{} could be adversely affected. \figref{fig:noise} demonstrates \spd{} curves for noisy quadratics generated from $y = x_1^2 + x_2 + 10 + \epsilon$ where $\epsilon \sim N(0,\sigma)$ $\epsilon$ and, at $\sigma=2$, 95\% of the noise falls within [0,4] (since $2\sigma = 4$), meaning that the signal-to-noise ratio is at best 1-to-1 for $x_1^2$ in $[-2,2]$. For zero-centered Gaussian noise and this data set, \spd{} appears resilient, though $\sigma=2$ does show considerable variation across runs.  The superfluous noise variable $x_3$ in \figref{fig:interactions} also did not confuse \spd.

\begin{figure}[!htbp]
\begin{center}
\includegraphics[scale=0.45]{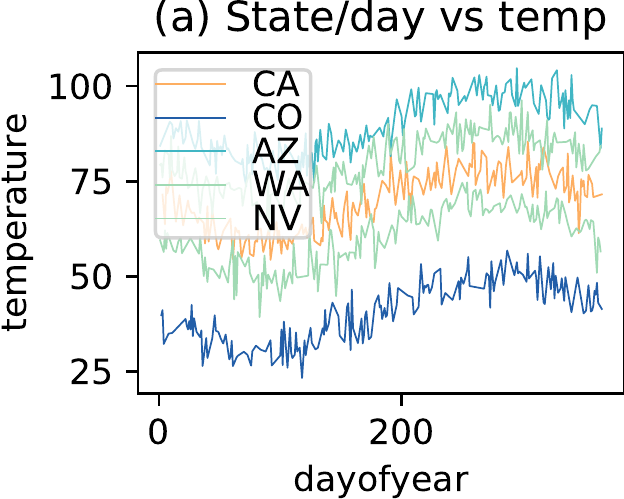}~~
\includegraphics[scale=0.45]{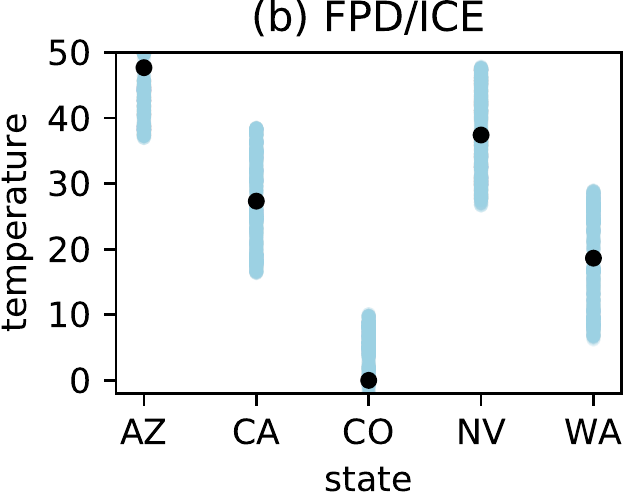}~~
\includegraphics[scale=0.45]{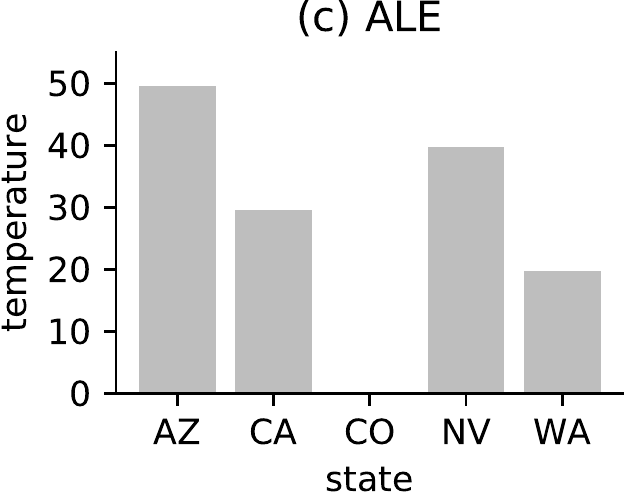}~~
\includegraphics[scale=0.45]{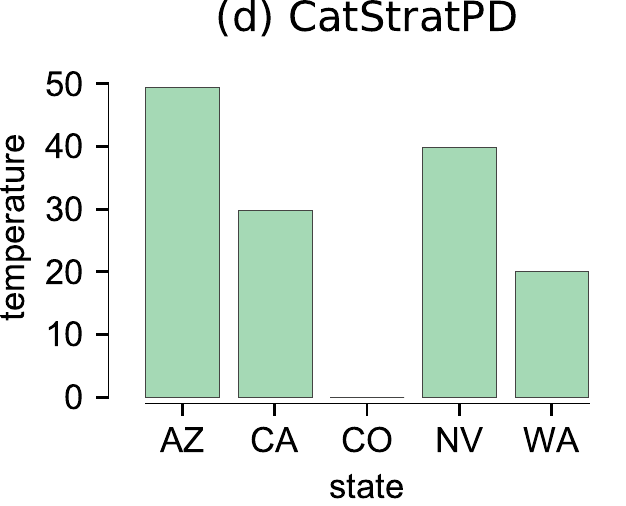}\vspace{-1mm}
\caption{\small $y = base[x_{\it state}] + 10 sin(\frac{2\pi}{365}x_{\it dayofyear}+\pi) + \epsilon$ where $\epsilon \sim N(\mu=0, \sigma=4)$. The {\em base} temperature per state is $\{{\it AZ}=90, {\it CA}=70, {\it CO}=40, {\it NV}=80, {\it WA}=60\}$. Sinusoids in (a) are the average of three years' temperature data.}
\label{fig:statetemp}
\end{center}
\end{figure}

Turning to categorical variables, \figref{fig:statetemp} presents partial dependence curves for FPD, ALE, and \cspd{} derived from a noisy synthetic weather data set, where temperature varies in sinusoidal fashion over the year and with different baseline temperatures per state. (The vertical ``smear'' in the FPD plot shows the complete sine waves but from the side, edge on.)  Variable {\tt\small state} is independent and all plots identify the baseline temperature per state correctly.

The primary goal of the stratification approach proposed in this paper is to obtain accurate partial dependence curves in the presence of codependent variables. To test \spd{}/\cspd{} and discover potential bias in existing techniques, we synthesized a body weight data set generated by the following equation with nontrivial codependencies between variables: 
\vspace{-1mm}

\begin{equation}\label{eq:weight}
{
\begin{array}{rll}
y & = &120 + 10(x_{height} - min(x_{height})) + 40x_{pregnant} - 1.5x_{education}\\[2pt]
\vspace{-10pt}\\[2pt]
\multicolumn{2}{r}{\text{where}} & x_{sex} \sim Bernoulli(\{M,F\}, p=0.5)\\[5pt]
                    & & x_{pregnant} = \begin{cases}
                                               Bernoulli(\{0,1\},p=0.5) & \text{ if } x_{sex} = F\\
                                               0 & \text{ if } x_{sex}=M\\
                                               \end{cases}\\[15pt]
                    & & x_{height} = \begin{cases}
                                               5*12+5+ \epsilon & \text{ if } x_{sex}=F,~ \epsilon \sim U(-4.5,5)\\	
                                               5*12+8 + \epsilon & \text{ if } x_{sex}=M,~ \epsilon \sim U(-7,8)\\
                                               \end{cases}\\[15pt]
                    & & x_{education} = \begin{cases}
                                               12 + \epsilon & \text{ if } x_{sex}=F,~ \epsilon \sim U(0,8)\\	
                                               10 + \epsilon & \text{ if } x_{sex}=M,~ \epsilon \sim U(0,8)\\
                                               \end{cases}
\end{array}
}
\end{equation}\vspace{1mm}

\begin{figure}[!htbp]
\begin{center}
\includegraphics[scale=0.55]{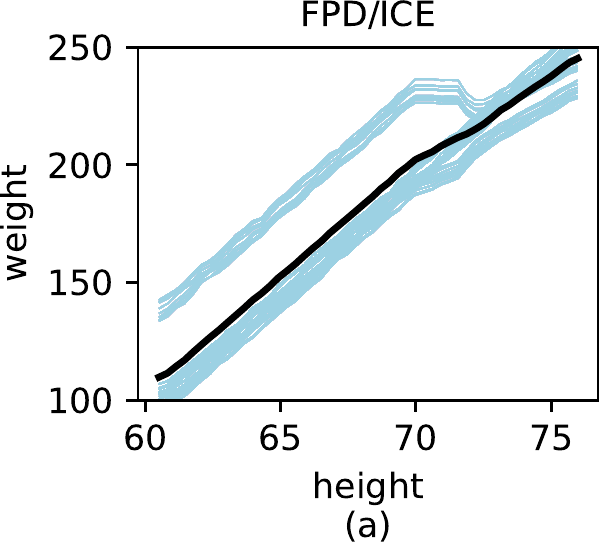}~\includegraphics[scale=0.55]{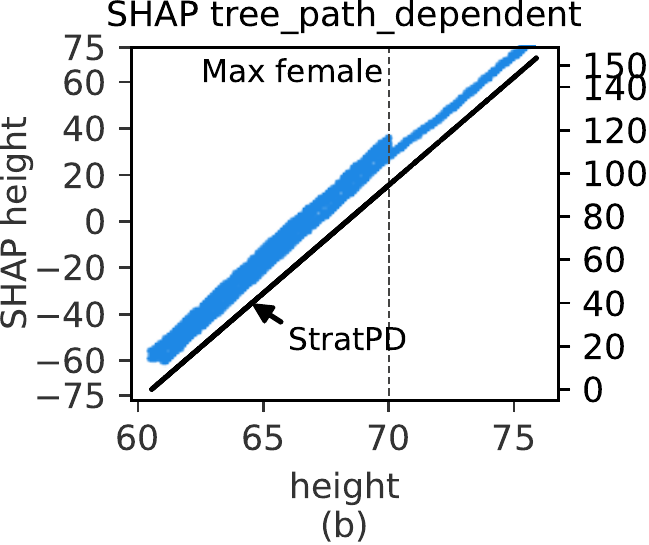}
\includegraphics[scale=0.55]{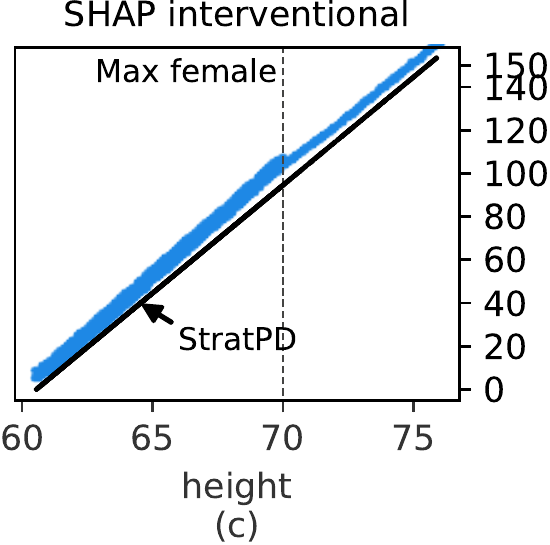}~
\includegraphics[scale=0.55]{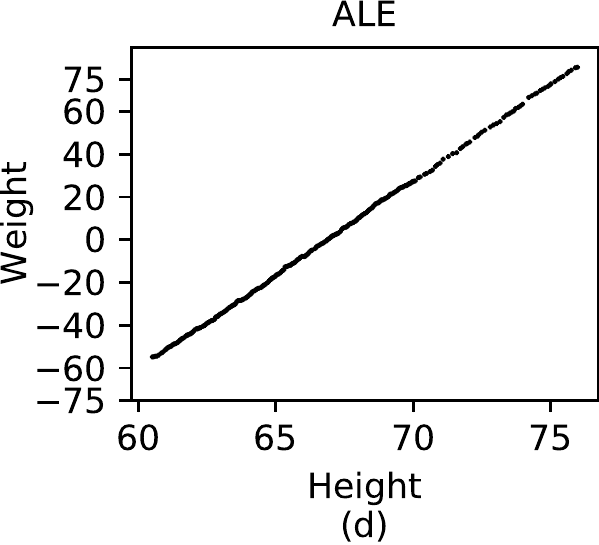}\vspace{-2mm}
\caption{\small Partial dependence plots of response body weight on feature $x_{height}$ using $n$=2000 synthetic observations from \eqref{eq:weight}. FPD and SHAP have ``kinks'' at the maximum female height. SHAP defines feature importance as the average SHAP value magnitude, which overemphasizes importance for heights below 70 inches here. The male/female ratio is 50/50, half of the women are pregnant, and pregnancy contributes 40 pounds. SHAP interrogated an RF tuned via 5-fold cross validation grid search (OOB $R^2$ 0.999) and explained all 2000 samples; the interventional case used 100 observations as background data. ALE used $K=300$.}
\label{fig:heightweight}
\end{center}
\end{figure}

\noindent The partial derivative of $y$ with respect to $x_{height}$ is 10 (holding all other variables constant), so the optimal partial dependence curve is a line with slope 10. \figref{fig:heightweight} illustrates the curves for the techniques under consideration, with ALE and \spd{} giving the sharpest representation of the linear relationship. (\spd's curve is drawn on top of the SHAP plots using the righthand scale.) The FPD and both SHAP plots also suggest a linear relationship, albeit with a little less precision. The ICE curves in \figref{fig:heightweight}(a) and ``fuzzy'' SHAP curves have the advantage that they alert users to variable  dependencies or interaction terms.  On the other hand, the kink in the partial dependence curve and other visual phenomena could confuse less experienced machine learning practitioners and certainly analysts and researchers in other fields (our primary target communities).

Even for experts, explaining this behavior requires some thought, and one must distinguish between model artifacts and interesting phenomena. The discontinuity at the maximum female height location arises partially from the model having trouble extrapolating for extremely tall pregnant women. Consider one of the upper ICE lines in \figref{fig:heightweight}(a) for a pregnant woman. As the ICE line slides $x_{height}$ above the maximum height for a woman, the model leaves the support of the training data and predicts a {\em lower} weight as height increases (there are no pregnant men in the training data). ALE's curve is straight because it focuses on local effects, demonstrating that the lack of sharp slope-10 lines for FPD and SHAP cannot be attributed simply to a poor choice of model.  

\begin{figure}[!htbp]
\begin{center}
\includegraphics[scale=0.48]{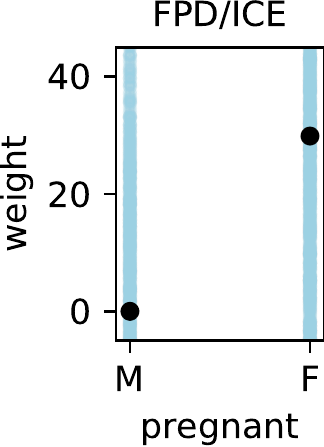}~~
\includegraphics[scale=0.48]{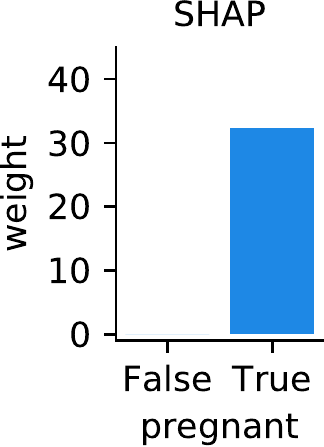}~~
\includegraphics[scale=0.48]{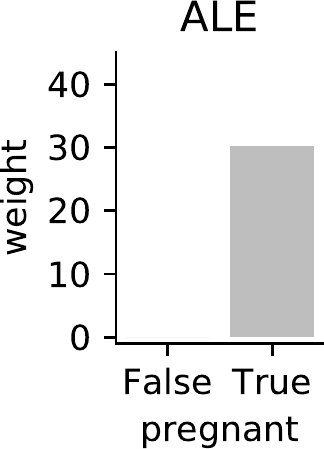}~~
\includegraphics[scale=0.48]{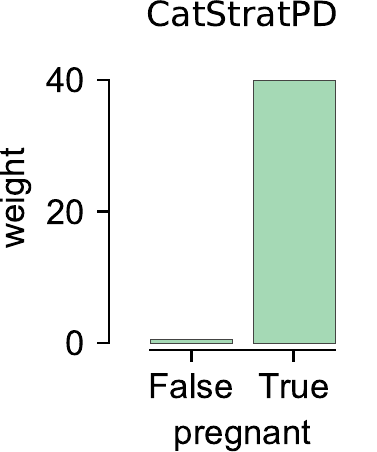}\vspace{-2mm}
\caption{\small  Partial dependence bar charts for boolean $x_{pregnant}$ with the same model from \figref{fig:heightweight}. Only \cspd{} gets the correct 40 pound contribution per \eqref{eq:weight}.}
\label{fig:pregnant}
\end{center}
\end{figure}

Also, SHAP defines $x_j$ feature importance as the average magnitude of the $x_j$ SHAP values, which introduces a paradox.  The spread of the SHAP values alerts users to variable interactions, but allows contributions from other variables to leak in, thus, potentially leading to less precise estimates of $x_{height}$'s importance. The average SHAP magnitude skews upward, in this case, because of the contributions from pregnant women.

It is conceivable that a more sophisticated model (in terms of extrapolation) could sharpen the FPD and SHAP curves for $x_{height}$. There is a difference, however, between extrapolating to a meaningful but unsupported vector and making predictions for nonsensical vectors arising from variable codependencies.  Techniques that rely on such predictions make the implicit assumption of variable independence, introducing the potential for bias. Consider \figref{fig:pregnant} that presents the partial dependence results for categorical variable $x_{pregnant}$ (same data set). The weight gain from pregnancy is 40 pounds per \eqref{eq:weight}, but only \cspd{} identifies that exact relationship; FPD, SHAP, and ALE show a gain of 30 pounds. 

\cspd{} stratifies persons with the same or similar sex, education, and height into groups and then examines the relationship between $x_{pregnant}$ and $y$. If a group contains both pregnant and nonpregnant females, the difference in weight will be 40 pounds in this noiseless data set (if we assume identical \xnj).  FPD and ALE rely on computations that require fitted models to conjure up predictions for nonsensical records representing pregnant males (e.g., $\hat{f}(x_j = pregnant, {\bf X}_{\slash j})$). Not even a human knows how to estimate the weight of a pregnant male. SHAP, per its definition, does not require such predictions, but in practice for efficiency reasons, SHAP approximates $\Ex[\hat{f}(x_{j} = pregnant,{\bf X}_{\slash j}) | {\bf X}_j = x_j]$ with $\Ex[\hat{f}(x_{j} = pregnant,{\bf X}_{\slash j})]$, which does not restrict pregnancy to females. As discussed above, there are advantages to all of these model-based techniques, but this example demonstrates there is potential for partial dependence bias.

\cut{
pregnant female at max range [233.33467678]
pregnant female in male height range [228.71359869]
nonpregnant female in male height range [209.88006493]
male in male height range [219.46772083]
}

\cut{
$1/n*\sum_{i=1}^n f(x_j = pregnant, x_{i, \bar{j}}) - f(x_j = not pregnant, x_{i, \bar{j}})$

\noindent where ${x_{i, \bar{j}}: i=1,2,?,n}$ are the n training observations of $x_{\bar{j}}$
}

\begin{figure}[!htbp]
\begin{center}
\includegraphics[scale=0.35]{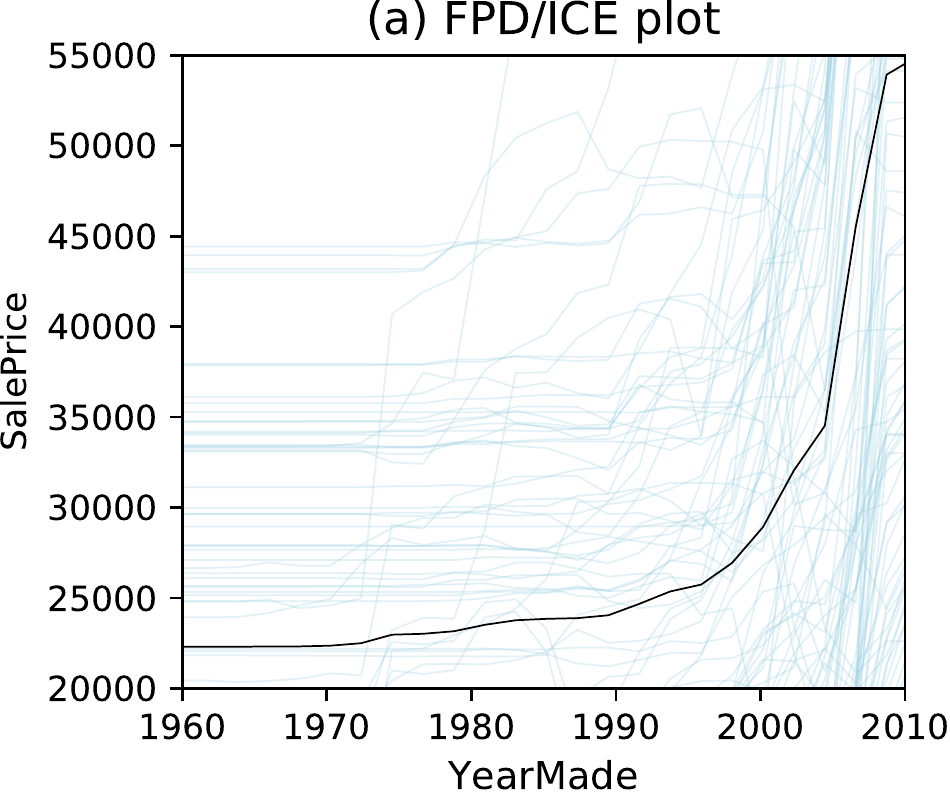}~~
\includegraphics[scale=0.35]{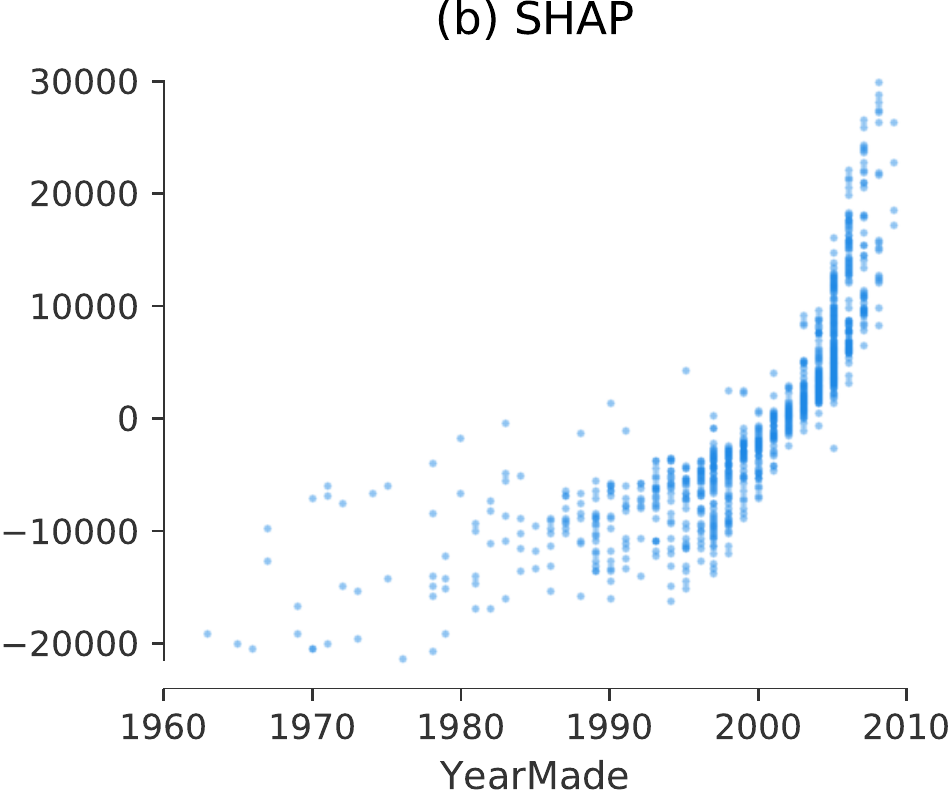}~~
\includegraphics[scale=0.35]{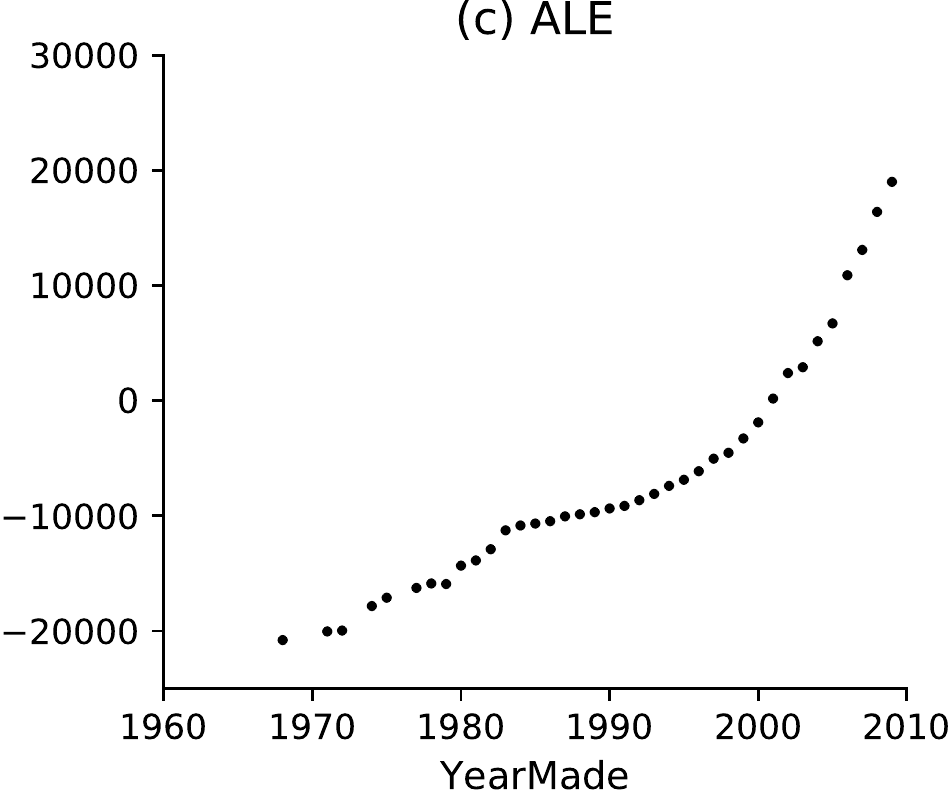}~~
\includegraphics[scale=0.35]{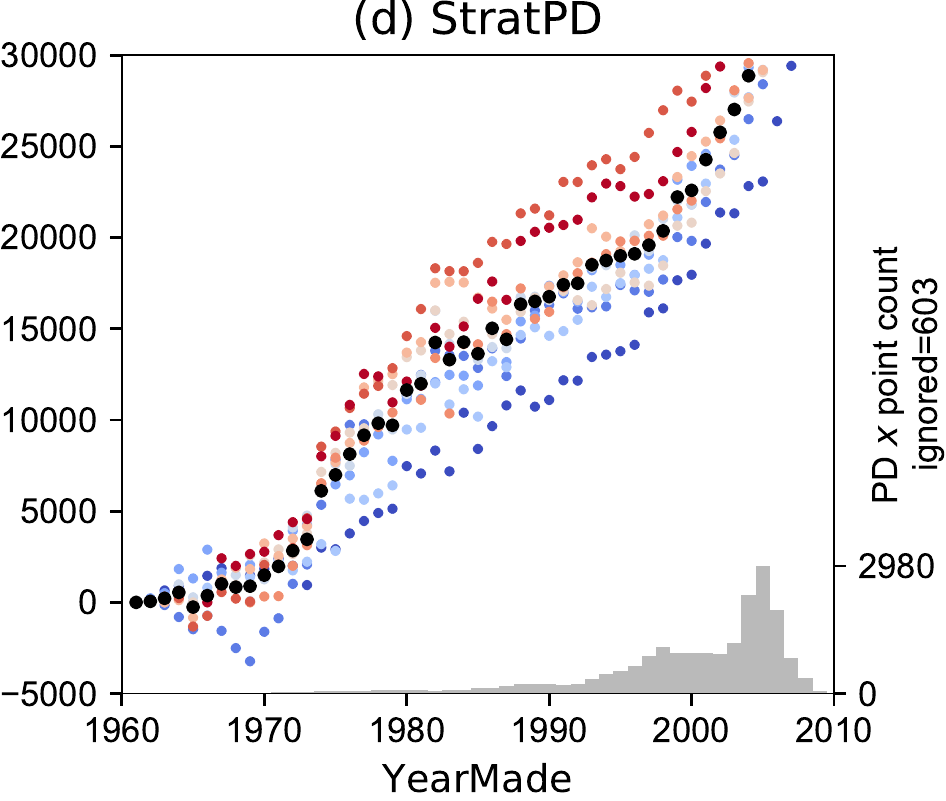}\vspace{-2mm}
\caption{\small Partial dependence curves of bulldozer {\tt YearMade} versus {\tt SalePrice} for FPD/ SHAP, ALE, and \spd. $n$=20,000 observations drawn from \textasciitilde{}362k. SHAP interrogated a random forest (OOB $R^2=0.85$) to explain 1000 training observations with 100 observations as background data.  ALE used $K=300$. Hyper parameters were tuned using 5-fold cross validation grid search over several hyper parameters.}
\label{fig:yearmade}
\end{center}
\end{figure}

The stratification approach also gives plausible results for real data sets, such as the bulldozer auction data from \citet{bulldozer}. \figref{fig:yearmade} shows the partial dependence curves for the same set of techniques as before on feature {\tt\small YearMade}, chosen as a representative because it is very predictive of sale price. The shape and magnitude of the FPD, SHAP, ALE, and \spd{} curves are similar, indicating that older bulldozers are worth less at auction, which is plausible. The \spd{} curve shows 10 bootstrapped trials where the heavy black dots represent the partial dependence curve and the other colored curves describe the variability. 

\begin{figure}[!htbp]
\begin{center}
\includegraphics[scale=0.35]{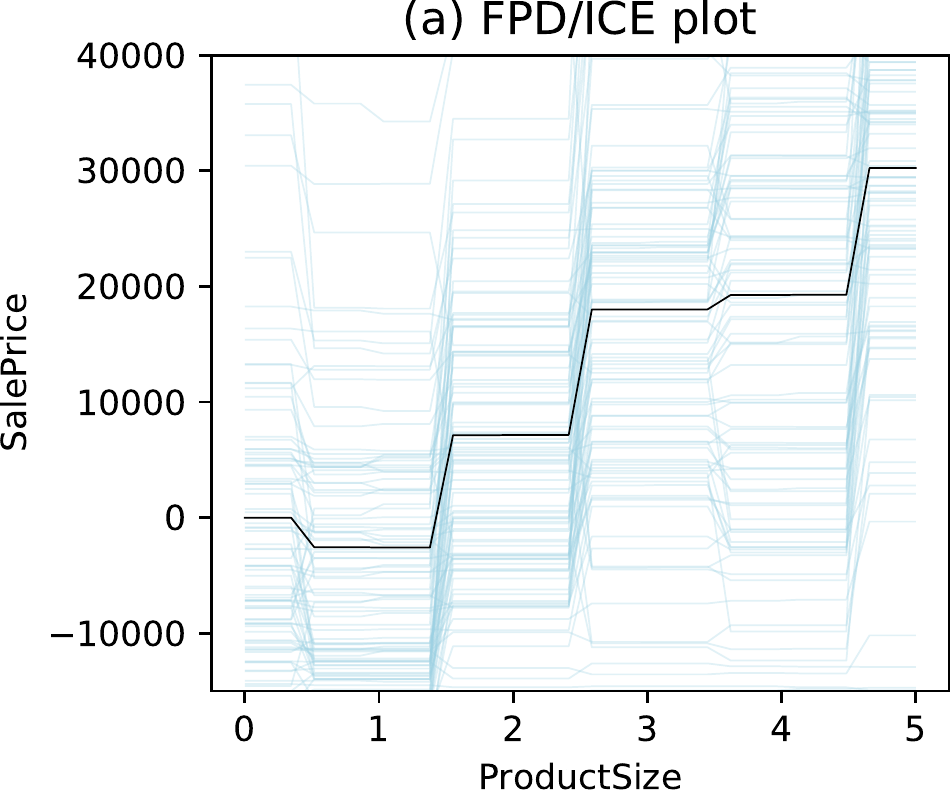}~~
\includegraphics[scale=0.35]{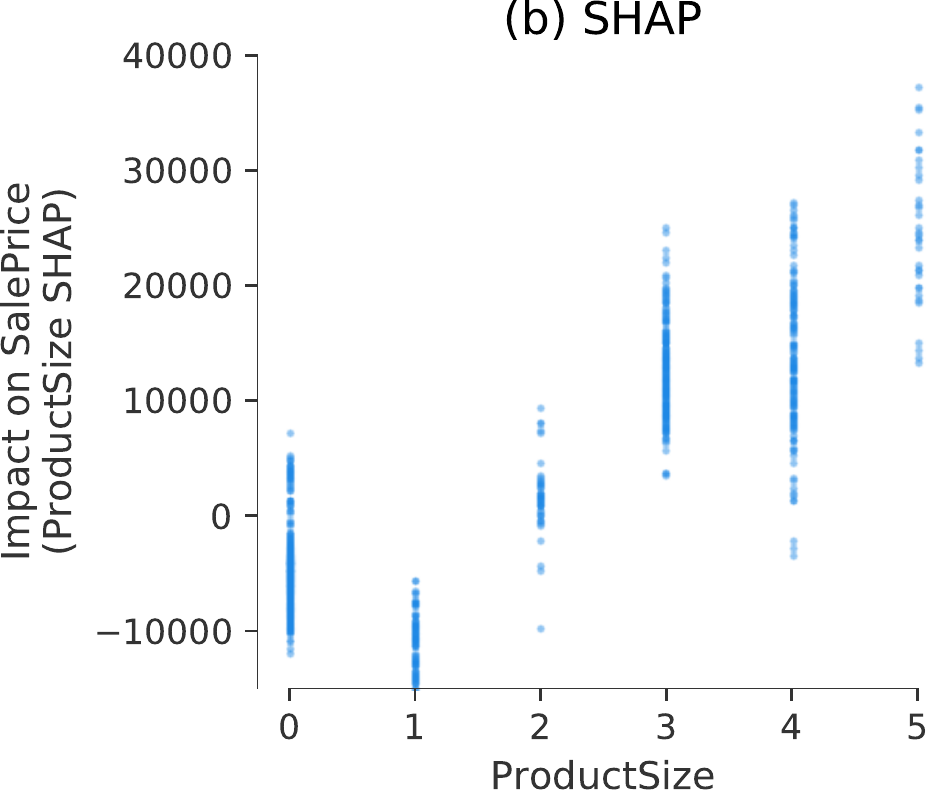}~~
\includegraphics[scale=0.35]{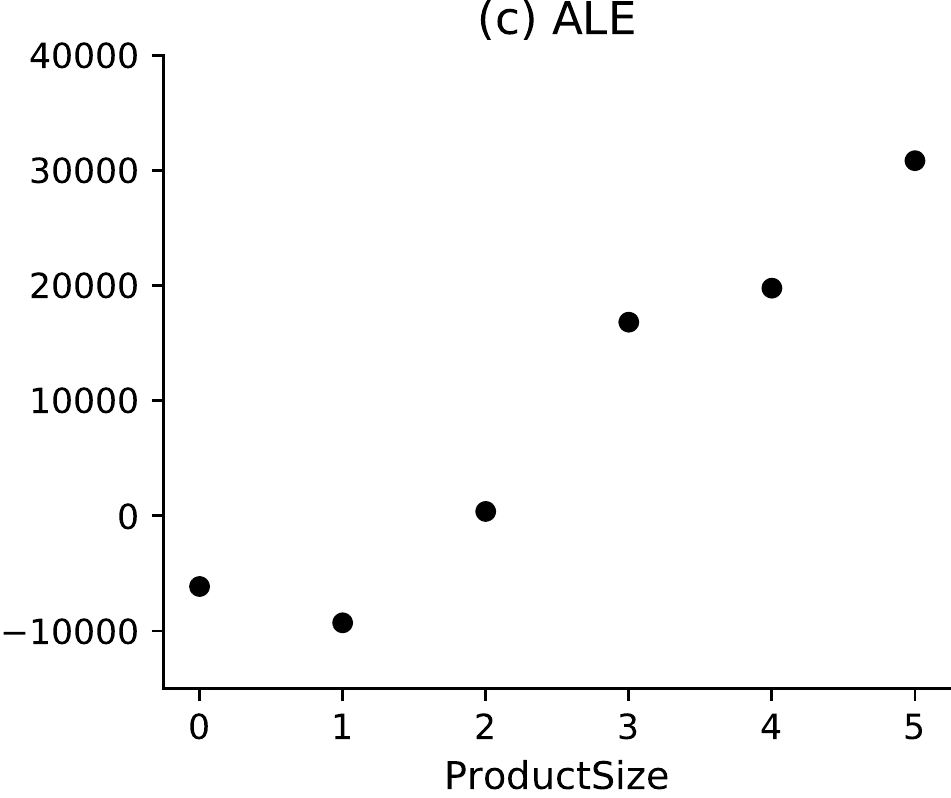}~~
\includegraphics[scale=0.35]{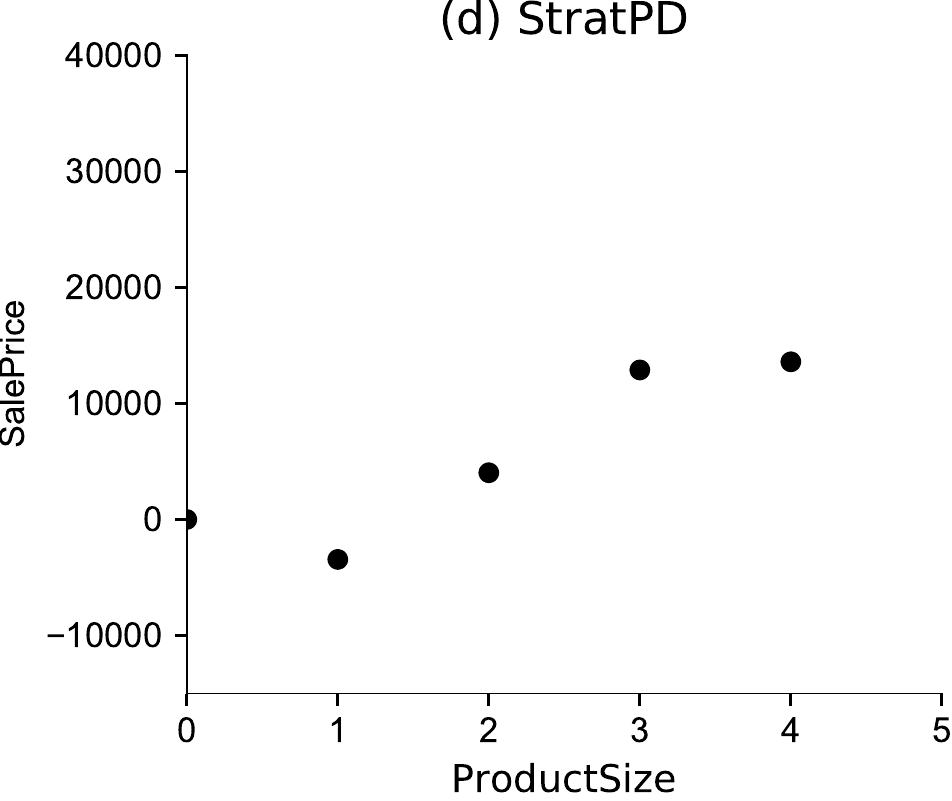}
\caption{\small Partial dependence curves of bulldozer {\tt ProductSize} versus {\tt SalePrice} for FPD/ICE, SHAP, ALE, and \spd. $n$=20,000 observations drawn from \textasciitilde{}362k. Same model, setup as in \figref{fig:yearmade}.  The \spd{} dots are the average of 10 bootstrapped trials and the partial dependence for {\tt ProductSize} is missing because a forward difference is unavailable at the right edge.}
\label{fig:ProductSize}
\end{center}
\end{figure}

As a second example, consider the curves for feature {\tt\small ProductSize} in \figref{fig:ProductSize}. All plots have roughly the same shape but the \spd{} plot lacks a dot for {\tt\small ProductSize}=5 because  forward differences are unavailable at the right edge.  (We anticipate switching switching to a central difference to avoid this issue with low-cardinality discrete variables.) It also appears that \spd{}  considers {\tt\small ProductSize}'s 3 and 4 to be worth less than suggested by the other techniques, though \spd{}'s might be in line with the average SHAP plot values.

\begin{figure}[!htbp]
\begin{center}
\includegraphics[scale=.45]{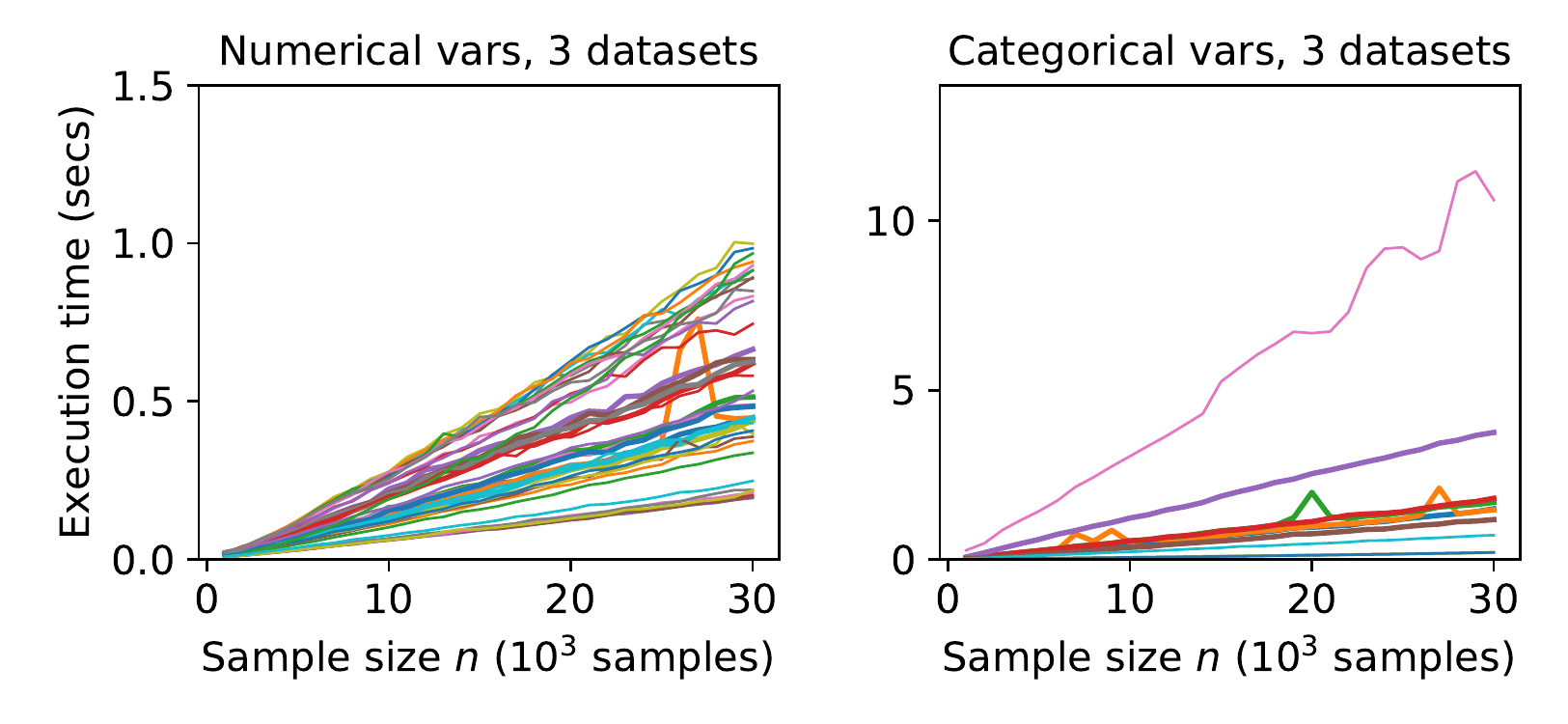}\vspace{-3mm}
\caption{\small Time to compute partial dependence curve for up to 30,000 observations for 40 numerical and 11 categorical variables. $p=20$, bulldozer $p=14$, and flight $p=17$. The Numba just-in-time compiler is used to improve performance, but timing does not include compilation; users do experience this ``warm-up'' time.}
\label{fig:timing}
\end{center}
\end{figure}

And, finally, an important consideration for any tool is performance, so we plotted execution time versus data size (up to 30,000 observations) for three real Kaggle data sets: rent, bulldozer, and flight arrival delays \citep{flights}. \figref{fig:timing} shows growth curves for 40 numerical variables and 11 categorical variables grouped by type of variable.  For these data sets, \spd{} takes 1.2s or less to process 30,000 records for any  numerical $x_j$, despite the potential for quadratic cost. \cspd{} typically processes categorical variables in less than 2s but takes 13s for the high-cardinality categorical {\tt\small ModelID} of bulldozer (which looks mildly quadratic).  These elapsed times for our prototype show it to be practical and competitive with FPD/ICE, SHAP, and ALE.  If the cost to train a model using (cross validated) grid search for hyper parameter tuning is included, \spd{} and \cspd{} outperform these existing techniques (as training and tuning is often measured in minutes).

\cut{
flight shape (5714008, 17), 6 cats
rent shape (49352, 20), no cats
bulldozer shape (362781, 14) records, 5 cats
}

\section{Conclusion and future work}

In this paper, we contribute a method for computing partial dependence curves, for both numerical and categorical explanatory variables, that does not use  predictions from a fitted model.   Working directly from the data  makes partial dependences accessible to business analysts and scientists not qualified to choose, tune, and assess machine learning models.  For experts, it can  provide hints about the relationships in the data to help guide their choice of model. Our experiments show that \spd{} and \cspd{} are fast enough for practical use and correctly identify partial dependences for synthetic data and give plausible curves on real data sets. \spd{} relies on two important hyper parameters (with broadly applicable defaults) but model-based techniques should include the hyper parameters of the required fitted model for a fair comparison.  Our goal here is not to  argue  that model-based techniques are not useful. Rather, we are pointing out potential issues and hoping to open a new line of nonparametric inquiry that experiments have shown to be applicable in situations and accurate in cases where model-based techniques are not. An obvious next goal for this approach is a version for classifiers and to extend the technique to two variables, paralleling ALE's second order derivative approach. 

{
\bibliography{stratpd}
}


\section{Appendix}

Pseudocode for \spd{} and \cspd.\\

\SetAlgoNoEnd%
\setlength{\algomargin}{5pt}
\begin{algorithm}[H]
\SetAlgoLined
\DontPrintSemicolon
\SetAlgorithmName{Algorithm}{List of Algorithms}
\SetAlgoSkip{}
\TitleOfAlgo{{\em StratPD}({\bf X}, {\bf y}, $j$, {\it min\_samples\_leaf}, {\it min\_slopes\_per\_x})}
$T$ := Decision tree regressor fit to (\Xnj{}, $\bf y$) with hyper parameter: ${\it min\_samples\_leaf}$\;
\For{each leaf $L \in T$}{
        ${\bf x}_L := \{x_j^{(i)}\}_{i \in L}$, the unique and ordered $x_j$ in $L$\Comment*{\it Get leaf observations}
        $\bar{\bf y}_L^{(k)} := \Ex[ y^{(i)} \,|\, (x_j^{(i)}=x_L^{(k)},  y^{(i)})] \text{ for } k = 1..|{\bf x}_L|, i \in L$\Comment*{\it Compute $\bar{y}$ per unique $x_j$ value}
        $\delta_L^{(k)} = \frac{\bar{\bf y}_L^{(k+1)} - \bar{\bf y}_L^{(k)}}{{\bf x}_L^{(k+1)} - {\bf x}_L^{(k)}}$\Comment*{\it Discrete forward difference between adjacent unique $x_j$ values}
	Add tuples ${({\bf x}_L^{(k)}, {\bf x}_L^{(k+1)},~ \delta_L^{(k)})}$  to list ${\bf D}$ for $k = 1..|{\bf x}_L|-1$\\
}
${\bf ux} := \{x_j^{(i)}\}_{i \in 1..n}$, the unique and ordered $x_j$ in ${\bf X}_j$\\
Let $\bf c$ and $\delta$ be vectors of length $|\bf ux|$\\
\For(\hfill$\triangleright$\ {\it\textcolor{gray}{\small Count slopes and compute average slope per unique $x_j$ value}}){each $x \in {\bf ux}$}{
	${\bf slopes}_x$ := [$slope$ for $(a,b,slope) \in \bf D$ if $x \ge a$ and $x <b$]\\
	${\bf c}_x$ := $|{\bf slopes}_x|$\Comment*{\it Count number of slopes used to estimate overall slope at $x$}
	${\delta}_x$ := $\overline{\bf slopes}_x$\Comment*{\it Slope estimate at $x$ is average of slopes whose range overlaps $x$}
}
${\bf \delta}$ := ${\bf \delta}[{\bf c} \ge min\_slopes\_per\_x]$\Comment*{\it Drop missing slopes and those computed with too few}
${\bf ux}$ := ${\bf ux}[{\bf c} \ge min\_slopes\_per\_x]$\Comment*{\it Drop $x_j$ values without sufficient estimates}
${\bf pd}_x$ := ${\bf ux}^{(k+1)} - {\bf ux}^{(k)}$ for $k = 1..|{\bf ux}|-1$\\
${\bf pd}_y$ := [0] + cumulative\_sum($\delta \times {\bf pd}_x$)~~~\Comment*{\it integrate, inserting 0 for leftmost $x_j$}
\Return{${\bf pd}_x, {\bf pd}_y$}
\label{alg:StratPD}
\end{algorithm}

\vspace{10mm}

\setlength{\algomargin}{5pt}
\begin{algorithm}[H]
\SetAlgoLined
\DontPrintSemicolon
\SetAlgorithmName{Algorithm}{List of Algorithms}
\SetAlgoSkip{}
\TitleOfAlgo{{\em CatStratPD}(${\bf X}, {\bf y}, j, {\it min\_samples\_leaf}$)}

\cut{
\KwOut{$\begin{array}[t]{l}
\Delta^{(k)} = \text{category } k \text{'s effect on } y \text{ where } mean(\Delta^{(k)})=0\\
n^{(k)} = \text{number of supported observations per category $k$}\\
\end{array}$
}
}
$n_{cats} := |\{x_j^{(i)}\}_{i \in 1..n}|$, $n_{leaves} := |T|$\\
$T$ := Decision tree regressor fit to (\Xnj{}, $\bf y$) with hyper parameter: ${\it min\_samples\_leaf}$\;
Let $\Delta Y$ be a $n_{cats} \times n_{leaves}$ matrix whose columns, $\Delta Y_L$, are vectors of leaf category deltas\\
Let $C$ be a $n_{cats} \times n_{leaves}$ matrix whose columns, $C_L$, are vectors for leaf category counts\\
\For(\hfill$\triangleright$\ {\it\textcolor{gray}{\small Get average $y$ delta relative to random ref category for obs. in leaves}}){each leaf $L \in T$}{
        ${\bf x}_L := \{x_j^{(i)}\}_{i \in L}$, the unique $x_j$ categories in $L$\Comment*{\it Get leaf observations}
        $\bar{\bf y}_L^{(k)} := \Ex[ y^{(i)} \,|\, (x_j^{(i)}=x_L^{(k)},  y^{(i)})] \text{ for } k = 1..|{\bf x}_L|, i \in L$\Comment*{\it Compute $\bar{y}$ per unique $x_j$ category level}
        $C_L^{(k)} := |i \in L : x_j^{(i)}={\bf x}_L^{(k)}|$\Comment*{\it Count occurrences of each category}
	${\it refcat}$ := random category chosen from ${\bf x}_L$\\
	$\Delta {Y}_L$ = $\bar{\bf y}_L - \bar{\bf y}_L^{\it refcat}$\\
}
$\Delta {\bf y}$, {\bf c} := $\Delta {Y}_1$, $C_1$\Comment*{\it $\Delta {\bf y}$ is running average vector mapping category to average $y$ delta}
$completed$ := $\{L_1\}$; $work$ := $\{L_2 .. L_{n_{leaves}}\}$; \Comment*{\it sets of leaves}
\While(\hfill$\triangleright$\ {\it\textcolor{gray}{\small 2 passes is typical to merge all $\Delta {Y}_L$ into $\Delta {\bf y}$}}){$|work| > 0$ and $|completed|>0$}{
    completed := $\emptyset$\\
    \For{each leaf $L$ in work}{
        Let $common$ := categories in common between $\Delta {Y}_L$ and $\Delta \bf y$\\
        \If{$common \ne \emptyset$}{
            ${\it completed}$ := ${\it completed} \cup \{L\}$\\
            $cat$ := random category in $common$\\
            $\Delta Y_L$ := $\Delta {Y}_L- \Delta {Y}_{L}^{(cat)} + \Delta {\bf y}^{(cat)}$\Comment*{\it Adjust so $\Delta {Y}_{L}^{(cat)}=0$ then add corresponding $\Delta {\bf y}^{(cat)}$ value}
            $\Delta {\bf y}$ := $({\bf c} \times \Delta {\bf y} + C_L \times \Delta Y_L)/({\bf c} + C_L)$ where $z+NaN$={\it z}\Comment*{\it weighted average}
            ${\bf c} := {\bf c} + C_L$\Comment*{\it update running weight}
        }
    }
    $work := work \slash {\it completed}$\\
}
\Return{$\Delta {\bf y}$}\\
\label{alg:CatStratPD}
\end{algorithm}

\end{document}